\documentclass[10pt,letterpaper]{article}

\usepackage{iccm}
\usepackage{hyperref}
\iccmfinalcopy 

\usepackage{pslatex}
\usepackage{apacite}
\usepackage{float} 

\usepackage{amsmath}
\usepackage{amssymb}
\usepackage{mathtools}
\usepackage{bussproofs}
\usepackage[shortlabels]{enumitem}
\newcommand{\ostar}{~\textcircled{$\star$}~}

\newtheorem{definition}{Definition}

\title{Hey Pentti, We Did It Again!: \\ Differentiable vector-symbolic types that prove polynomial termination}
 
\author{{\large \bf Eilene Tomkins-Flanagan (eilenetomkinsflanaga@cmail.carleton.ca)}\\
        {\large \bf Connor Hanley (connorhanley@cmail.carleton.ca)}\\
                    {\large \bf Mary Alexandria Kelly (mary.kelly4@carleton.ca)} \\
  Department of Cognitive Science, Carleton University \\
  1125 Colonel By Drive, Ottawa, ON, K1S 5B6, Canada}

\begin{document}

\maketitle

\begin{abstract}
We present a typed computer language, Doug, in which all typed programs may be proved to halt in polynomial time, encoded in a vector-symbolic architecture (VSA). Doug is just an encoding of the light linear functional programming language (LLFPL) described by \citeA[ch. 7]{Schimanski2009}. The types of Doug are encoded using a slot-value encoding scheme based on holographic declarative memory (HDM; \citeNP{Kelly2020}). The terms of Doug are encoded using a variant of the Lisp VSA defined by \citeA{Flanagan2024}. Doug allows for some points on the embedding space of a neural network to be interpreted as types, where the types of nearby points are similar both in structure and content. Types in Doug are therefore learnable by a neural network. Following \citeA{Chollet2019}, \citeA{Card1983}, and \citeA{Newell1981}, we view \emph{skill} as the application of a procedure, or program of action, that causes a goal to be satisfied. Skill acquisition may therefore be expressed as \emph{program synthesis}. Using Doug, we hope to describe a form of learning of skilled behaviour that follows a human-like pace of skill acquisition \cite<i.e., \emph{substantially faster than brute force};>{Heathcote2000}, exceeding the efficiency of all currently existing approaches \cite{Kaplan2020,Jones2021,Chollet2024-Blog}. Our approach brings us one step closer to modeling human mental representations, as they must actually exist in the brain, and those representations' acquisition, as they are actually learned.

\end{abstract}


\citeA{Chollet2019} proposed a novel model for intelligence: an \emph{intelligent agent} does not just \emph{solve problems} in its environment in service to its goals (as we might infer it does in the Markov decision process model found in \emph{reinforcement learning}, \citeNP{Sutton2018}, pp. 46-53). Instead, an intelligent agent \emph{acquires the skills necessary to the task}, and uses its \emph{acquired skills} to solve problems in service to its goals. Chollet's model of ``skill'' is an extremely general one. In his view, a skill is a \emph{program}: a procedure made up of well-defined steps, which, when followed, will transform the environment from an initial state into a desired state. By separating the intelligent agent into an \emph{intelligent system} that generates skills, and \emph{skill programs} that realize the agent's competencies, Chollet frames the problem of intelligence as one of an ability to acquire skills from sparse examples in a great variety of problem types. Following Chollet, a \emph{greater intelligence} within some scope of problems is one that can acquire the skills to solve problems in that scope with a minimum of foreknowledge and training, and a \emph{more general intelligence} is intelligent, up to some degree, in a greater variety of scopes, more exhaustive of the space of possible problems.

The Chollet model may have a familiar ring for cognitive modelers, however. The \emph{Goals, Operators, Methods, and Selection rules} (GOMS; \citeNP{Card1983}, ch. 5) model describes skilled behaviour as a deployment of procedures that will tend to satisfy an agent's goals. In GOMS, an agent with some \emph{goal} $G$ uses its \emph{selection rules} to choose a \emph{method} that, the agent expects, will cause $G$ to be satisfied, if followed. A \emph{method}  is a procedure in the sense we used above: a sequence of well-defined steps. In the case of GOMS, the steps are \emph{operators}, discrete actions the agent can take that transform the state of the environment, or the agent's internal state. Following the procedure laid out by the selected method, the environment is transformed into a state whereby the goal $G$ is satisfied. \citeA{Newell1990} situates GOMS within the \emph{Soar} cognitive architecture, stating that GOMS ``posits a set of mechanisms and structures'' (p. 285) that can be formalized with the Soar architecture, and goes on to express skill acquisition in Soar (ch. 6.5) as a kind of search for increasingly efficient procedures that satisfy goals.


Chollet refers to the problem of acquiring skills as ``program synthesis'' (p. 30), that is, search for a procedure, with some given effect. \citeA[ch. 1.5]{Hutter2005} showed that synthesizing an optimal program that \emph{predicts} or \emph{represents} the environment allows an agent to behave in an optimally goal-directed (i.e., rational) way, with minimal extra machinery. In this way, the problems of \emph{skill acquisition} and \emph{representation learning} can be thought of as transformable into one another. 

Hutter, unfortunately, also finds (ch. 1.7), that program synthesis is profoundly nontrivial. In fact, finding a program that optimally predicts an arbitrary environment is incomputable, because it involves comparing programs in light of their computational effects, and some programs cannot be proven either to eventually halt or run forever \cite{Turing1936}, so their computational effects cannot be known either until they halt or until the end of time, whichever comes first. Even if we restrict ourselves to considering only programs up to some maximum finite length, if we are to be certain we have found the optimal representation, we must do an ugly exhaustive search through all possible programs up to our maximum length (which is intractable, in the sense of taking an exponentially long amount of time as a function of the maximum length) and, for each program, find its shortest representation that halts within some maximum time (which is also intractable). Hutter's finding of a doubly-exponential rate of skill acquisition is consistent with the ``power law'' described by \citeA{Newell1981}, however, \citeA{Heathcote2000} show that their law is an artifact of population averaging, and humans in fact acquire skill \emph{exponentially faster} than \citeauthor{Newell1981} believed. Unsurprisingly so, as \citeauthor{Newell1981} modeled skill acquisition as a simple, heuristic-informed (but in the worst case, like Hutter's model, brute-force) search, and managed to satisfy their power law. It seems obvious that, in order to be useful, a method of program synthesis that exhibits intelligence should be substantially better than this sort of brute-force approach.

\citeA{Chollet2019}'s ARC task was designed to demonstrate the efficiency of human intelligence relative to existing AI approaches, and the importance of efficient program synthesis. A model approaching the ARC task that showed some early success was DreamCoder \cite{Ellis2021}, which used a similar method to Hutter's, albeit synthesizing programs using a form of heuristic informed search involving a neural network. DreamCoder has since been surpassed by the GPT o-series models \cite{Chollet2024-Blog}, but \citeA{Chollet2025-StreetTalk} notes that both approaches seem to behave like exhaustive search:

\begin{quote}
It's always possible to ... logarithmically improve your performance by throwing more compute at the problem. And of course this is true for o1, but even before that it was also true for brute force program search systems. Assuming you have the right [domain-specific language], then extremely crude, basic, brute-force program iteration can solve ARC at human level. (49:55)
\end{quote}

We would like to constrain program synthesis so that search is restricted only to programs likely to be useful to solving a given problem. If we make good assumptions about the set of programs that are to be searched over, then, for many problems, searching through the constrained set of possible programs for their solution should become tractable. As a first pass, we should prohibit from consideration intractable programs, as we do not want to bother trying to evaluate them to discover their computational effects. This prohibition is achieved using a polynomial time type system \cite{girard_1998}. 

A \textit{type system} is a programming language that assigns, to each expression in a program, a \emph{type} that describes what kind of data it is, if it is a variable, and what it operates on and produces, if it is a function, accompanied with a set of \textit{inference rules} for how to demonstrate that some expression has some type. A \textit{polynomial time type system} is a type system that prohibits programs which do not have polynomial time complexity. A polynomial time type system is therefore not Turing-complete, but the set of problems polynomial time type systems are capable of representing still includes essentially all useful programs. The type system we will be considering is the \emph{Light Linear Functional Programming Language} \cite<LLFPL;>[ch. 7]{Schimanski2009}.

How can we make use of a polynomial time type system to constrain program synthesis? LLFPL, as an example, encodes the maximum recursion depth of a function in its type. We envision that types should be learnable, such that a learning agent should be able to acquire the ability to guess at the sort of structure a skill program should have, before it sets to work determining the program's specifics. The agent may need to revise its guess, but it should at least be capable of acquiring the ability to make good guesses. When a learning agent makes good guesses as to the structure of the program it should be synthesizing, its search is immediately restricted to just the programs of the appropriate structure. This set may still be quite large, but, if the appropriate structure restricts recursion depth, it is definitely much smaller than the set of all programs, or even all well-typed programs in the type system. 

Our goal, then, is to make LLFPL \emph{learnable by a neural network}. To do so, we will need to make it possible to express types as \emph{points in a vector space}, and we should make it so that structurally similar types are nearby in the space, such that a small change in position in the space connotes a small change in the structure of the type at the new position. In a word, we would like our types to be \emph{differentiable}.

Any arbitrary syntactic structure (of which types are a sort) can be encoded over a vector space in this way \cite{Flanagan2024}, using a \emph{vector-symbolic architecture} (VSA). 
In a VSA, we expect structures composed of distinct elements to be nearby in the vector space if (1) they have the same structure, and (2) the elements of which they are composed are also nearby. We will go beyond \citeauthor{Flanagan2024}, however, and take proper advantage of the features of a VSA to make distinct but similar structures spatially nearer to one another. We define a language, based on \citeauthor{Flanagan2024}'s Lisp VSA and LLFPL, called the \emph{vector-symbolic Lisp representation of the light linear functional programming language} (VSLRLLFPL), or Doug\footnote{Our implementation may be found at \url{https://github.com/eilene-ftf/doug}}, as the other name is very long.

\section{A body of many organs}

Doug builds on the work of \citeA{Flanagan2024} in order to (1) allow points on a neural network's \emph{embedding space} to encode \emph{systematically} decodable types, (2) constrain possible programs typed by any point on the embedding space to include only those that halt in polynomial time, and (3) for nearby points to encode types that are both \emph{structurally similar} and \emph{comprised of similar elements}. Therefore, the surface induced by the embedding space and some loss function on decoded types will be differentiable, where traversing the surface by \emph{gradient descent} will cause relatively smooth changes in the structure and content of types encoded at nearby points, even if not all points are decodable. In other words, \emph{the structure and content of types in Doug is learnable by a neural network}.

In order to achieve the three preceding goals, we must first select a polynomial time type system; as above, we use LLFPL. We will then recapitulate \citeauthor{Flanagan2024}'s definition of a VSA for the benefit of the novice reader, supplementing the discussion with the additional elements necessary to encode Doug, including \citeA{kymn_computing_2023}'s \emph{residue numbers}. We will then select a VSA-based encoding scheme whereby \emph{syntactically} similar structures are encoded as \emph{spatially} similar vectors, if they are similar in content. The encoding scheme we choose is the one employed by \citeA{Kelly2020}'s \emph{holographic declarative memory} (HDM). The preceding parts will come together so that, in the next section, we can encode the \emph{types} of Doug using the slot-value encoding of HDM with residue numbers, and the \emph{terms} using a variant of \citeA{Flanagan2024}'s Lisp VSA.

\subsection{Lightly linear flipout}

In this section, we will introduce some of the history and motivations behind linear type systems,
polyonomial time type systems, as well our language of choice, the \emph{Light Linear Functional Programming Language} \cite{Schimanski2009}.

\subsubsection{History and intuitions}

A type system is a full language specification that constrains the kinds of expressions in the 
language based on the notion of a \emph{type}. Types introduce
primitives into the system which regiment the kinds of things that our language deals with
and puts constraints on what we can do with those things \cite{nederpelt_type_2014}. 

One constraint that we might like to put on our programming language is that to
constrain the amount of usages of some item. Consider the following: suppose we are 
writing a program for a resource-constrained old computer, and we want to ensure
both (a) efficient (i.e., as little as possible) usage of the limited amount of memory available, 
and (b) ensure that all memory that we allocate for the program, if we do so, is neatly ``put 
back'' in place as quickly as possible. Motivations like these inspired \citeA{girard_1987} to formulate 
\emph{Linear Logic} (LL) which captures the aforementioned goals. Simply put, it does this
by restricting the number of usages of items by requiring that they are used \emph{exactly}
once in programs. But we say that a type system is \emph{affine} if and only if
it requires that terms be used \emph{at most} once.

\subsubsection{LLFPL}

\citeA{Schimanski2009} is a systematic study of polynomial time type systems.
The language that we will be encoding here is \citeA{Schimanski2009}'s own contribution: \emph{Light Linear Functional Programming Language} (LLFPL$_!$), which extends the \textit{Linear Functional Programming Language} \cite<LFPL;>{hofmann_lfpl} by 
combining it with elements from \emph{Light Linear Logic} \cite<LLL;>{girard_1998}. Here we will lay out the language definition in the 
standard way, using Backus-Naur form.  
After, we will give a natural language explanation of what the constants are meant to denote.

\begin{definition}[Types of LLFPL; Levels of types]\label{def}
    The set of \emph{types} of LLFPL are defined by the following expression,
    \begin{equation}
        \sigma, \tau ~::=~B^n ~|~\sigma \multimap \tau ~|~\sigma \otimes \tau ~|~ L^n (\sigma) 
        ~|~ !^n \sigma ~|~ \diamond^n.
    \end{equation}
    The \emph{level} of a type is defined recursively by the function,
    \begin{equation}
        \ell (\rho) := \begin{cases}
            n ~\text{if}~\rho \in \{B^n, L^n(\sigma),!^n \sigma, \diamond^n\}, \\
            \min\{ \ell(\sigma), \ell(\tau) \}, ~\text{otherwise}.
        \end{cases}
    \end{equation}
\end{definition}
More naturally, $B^n$ is the type of \emph{Booleans}, $\sigma \multimap \tau$  the \emph{linear function} type. $\sigma \otimes \tau$ is the tuple type of pairs, $L^n(\sigma)$ is the type of 
lists of type $\sigma$.  $!^n \sigma$ is
a modal operator denoting that the embedded type $\sigma$ can be used ``arbitrarily
often'' \cite{Schimanski2009}. $\diamond^n$ is a \emph{credit} type from 
\citeA{hofmann_lfpl}, which is used to limit recursion depth.

\begin{definition}[The constants of LLFPL]\label{const}

    The \emph{constant} terms of LLFPL, which are \emph{constructors} and 
    \emph{destructors} for the types given in Definition~\autoref{def}.
    \begin{align}
        \mathbf{tt}^n, \mathbf{ff}^n &: B^n, \\
        \mathbf{Case}^n_\sigma &: B^n \multimap \sigma \multimap \sigma \multimap \sigma, \\
        \mathbf{Case}^n_{\tau, \sigma} &: L^n(\tau) \multimap (\diamond^n\multimap \tau \multimap L^n(\tau) \multimap \sigma) \multimap \sigma \multimap \sigma, \\
        \mathbf{cons}^n_\tau &: \diamond^n \multimap \tau \multimap L^n(\tau), \\
        \mathbf{nil}^n_\tau &: L^n (\tau), \\
        d^n &: \diamond^n, \\
        \otimes^n_{\tau, \rho} &: \tau \multimap \rho \multimap \tau \otimes \rho, \\
        \pi^n_\sigma &: \tau \otimes \rho \multimap (\tau \multimap \rho \multimap \sigma) \multimap \sigma.
    \end{align}
    Where for Eq.'ns (4-5), we have the constraint that $\ell (\sigma) \geq n$,  for Eq. (9)
    that $\ell (\sigma \otimes \tau) = n$, and finally for Eq. (10) that $\ell (\sigma) \geq \ell (\tau \otimes \rho) = n$.
\end{definition}
Here, the intuitive ``meaning'' behind each constant is listed: $\mathbf{tt}^n, \mathbf{ff}^n$ are
\emph{true} and \emph{false} constants, or $\top$ and $\bot$. $\mathbf{Case}^n_\sigma$ and 
$\mathbf{Case}^n_{\sigma, \tau}$ are \emph{destructors} for boolean types and list types
(respectively); $\mathbf{cons}^n_\tau$ and $\mathbf{nil}^n_\tau$ are the constructor
and base element of list types; $d^n$ is a \emph{chit} of the \emph{credit type}, a 
sort of token you have to give to recursive procedures that is consumed on usage in order
to limit the depth of recursion; $\otimes^n_{\tau, \rho}$ is the constructor for tuple types;
and finally, $\pi^n_\sigma$ is the \emph{projection function}, which is the destructor
for tuples types.

The language is not just a collection of types and constants, but also terms which 
form the object-level of the language. Terms are what are used to express the procedures
which an interpreter, computer, or agent, follows.
\begin{definition}[Terms of LLFPL]\label{terms}
    The \emph{terms} of LLFPL are defined inductively,
    \begin{align}
        s, t ::= x : \tau &~|~ c ~|~ (\lambda x : \tau . t) ~|~ (t~s) ~|~ !^n \boxed{x = \{s\}_{!^n}~\text{in}~t}\\
        & ~|~ !^n \boxed{t} ~|~\{t\}_{!^n} ~|~ \left(s \overset{n}{\triangleleft}^{x_1}_{x_2} t \right). \nonumber
    \end{align}
    where we have types $\tau$, constants $c$, natural numbers $n$, and variables $x_1, \ldots, x_m$. Constants of LLFPL are terms.
\end{definition}
More naturally, $x$'s are variables, $c$'s constants, $(\lambda x : \tau . t)$ is a 
$\lambda$-abstraction \cite[p. 2]{nederpelt_type_2014}, and $(t~s)$ an application of a
function $t$ to $s$. The special terms here are the \emph{boxed terms}
$!^n \boxed{\cdot}$. A boxed term is ``closed'' (i.e., has no free variables), except for those bound by terms in holes $\{\cdot\}_{!n}$ \cite[p. 200]{Schimanski2009}. A term $t$ enclosed by a box $!^n\boxed{t}$ has level $n+1$. \, When a hole is filled in, as in $!^n \boxed{x = \{s\}_{!^n}~\text{in}~t}$, we bind $x$ to the value $s$ in the term $t$, similarly to a function application. When a boxed term is enclosed in a hole of the same level $\{!^n\boxed{t}\}_{!n}$, the bang and box are eliminated, and $t$ is lowered from level $n+1$ to level $n$. 

A type system is not complete without accompanying inference rules, which are a collection
of rewriting rules specifying when and under what conditions terms can be
properly said to have some type. We will not be enumerating the inference rules of 
the type system. For a complete presentation of LLFPL's inference rules see \citeA[pp. 209-210]{Schimanski2009}

\subsubsection{How LLFPL is polynomial}

\citeA[ch. 7.3.5]{Schimanski2009} proves that LLFPL can only
express programs that halt in polynomial time. Many useful functions, like \emph{quicksort}, may be expressed in LLFPL, but intractable functions cannot be. LLFPL achieves its polynomial restriction by a careful interplay of the credit and boxed expressions during the evaluation of recursive expressions. The key evaluation rule to consider is that defined for \emph{folded} expressions, where a function mapped over a list has a variable bound to a holed term.

\begin{align}
    & \left((\mathbf{cons}_{n+1}^\tau t^{\diamond^{n+1}} v \ l) \ f[z := \{r\}_{!n}]\  g\right) \mapsto_{n+1}^l \\ 
    & \qquad \quad \left(r \overset{n}{\triangleleft}^{r_1}_{r_2} (f[z := \{r_1\}_{!n}] \ t \ v \ (l \ f[z := \{r_2\}_{!n}] \ g))\right) \nonumber
\end{align}

The left side of the above should be read as the \emph{application of a list} consisting of a head $v$ and a tail $l$ to a recursive case $f$ and a base case $g$, where the variable $z$ is bound to the holed value $r$ in $f$. When a list is ``applied'', it just means that $f$ is to be folded over each value of the list in turn, until the base case. On the right hand side, we have that the \textit{multiplexer} $\triangleleft$ copies $r$ into $r_1, r_2$, which must be done explicitly since copying is restricted. Then, $f$ is applied to $t$, the credit, $v$, the head, and the result of the recursive map over the tail $l$.

Because applying $f$ consumes a chit, which are stored in lists, there must be linearly many calls to $f$ in the length of the list. Iteration calls $f$ multiple times, but it can't make exponentially proliferating recursive calls. A variable in $f$ may be bound to another recursive term, allowing nested recursion, but since it's a holed term, it must be one level below $f$. As a result, linear recursive calls can be nested only up to the maximum level of a term. \textit{In order to increase the maximum polynomial order of a term, one must increase its level}.

\subsection{What's a VSA again?}

\citeA{Flanagan2024} define a VSA as:

\begin{quote}
A vector-symbolic architecture is an algebra (i.e., a vector space with a bilinear product),
\begin{enumerate}
    \item that is closed under the product $\otimes: V \times V \to V$ (i.e., if $u \otimes v = w$, then $u, v, w \in V$)
    \item whose product has an ``approximate inverse'' $\overline{\otimes}$ that, given a product $w$ and one of its operands $u$ or $v$, yields a vector correlated with the other operand
    \item for which there is a dogma for selecting vectors from the space to be treated as atomic ``symbols'' (yielding themselves, thereby, to syntactic manipulations defined in terms of the algebra), 
    \item that is paired with a memory system $\mathcal{M}$ that stores an inventory of known symbols for retrieval after lossy operations (e.g., involution), that can be recalled from $\mathcal{M}(p)$, and which is appendable $\mathcal{M} \twoheadleftarrow t$, and
    \item possesses a measure of the correlation (a.k.a., similarity) of two vectors, $\mathbf{sim}(u, v) \in [-1, 1]$, where $1$ and $-1$ imply that $u, v$ are colinear, $0$ that they are linearly independent.
\end{enumerate}
\end{quote}

Certain VSAs relax the above properties, but all behave in a manner that approximates these properties. \citeauthor{Flanagan2024} also show that VSAs are \emph{Cartesian closed} under these properties, meaning that a VSA can express an arbitrary \emph{Turing-complete language} over vectors of fixed dimension (so long as the memory may be arbitrarily large). For our convenience, we extend the definition of a generic VSA with \emph{permutations}, used in HDM, as well as a \emph{second} product operator $\star$ and \emph{resonator networks}, used in the \emph{residue numbers} we employ to encode the natural numbers.

A \emph{permutation} $\mathbf{P}_{c}(v)$ is a function that reorders the dimensions of a vector $v$. That is, for a finite vector $v \in V$, $\mathbf{P}_c(v) = v' \in V$ where $v'_j = v_i$, where all values of $v_i$ are mapped to exactly one $v'_j$. Permutations are invertible, so $\mathbf{P}_c^{-1}(\mathbf{P}_c(v)) = v$.  The \emph{second product} $\star$ behaves as $\otimes$, except that $a \star b \neq a\otimes b$ in general, and each product has a different multiplicative unit. Given some composite representation  $v = \bigotimes_{i=1}^k a_i$, where $a_i \in A_i$, a resonator network decomposes $v$ into a tuple of the elements of which it is composed: $R(v, A_1, ... A_k) = (a_1, ..., a_k)$.

\subsubsection{Permutative concerns} Permutations are typically applied together with the first product operator $\otimes$ in order to achieve \emph{asymmetric binding}. That is, where $a \ostar b = \mathbf{P}_{\mathrm{right}}(a) \otimes \mathbf{P}_{\mathrm{left}}(b)$ for constant permutations $\mathbf{P}_{\mathrm{left}}, \mathbf{P}_{\mathrm{right}}, \mathbf{P}_{\mathrm{left}} \neq \mathbf{P}_{\mathrm{right}}$, we have that $a \ostar b \neq b \ostar a$. This allows for the inductive encoding of sequences $a \ostar (b \ostar (c \ostar ...))$.

\subsubsection{Residual notes} Residue numbers use complex-domain holographic reduced representations \cite{Plate2003}. We will try to make the treatment as generic as possible for the purpose of our formalism. Given a function $\zeta$ that generates a VSA representation of a natural number $n$, $\zeta(n) = v$, we define the sum as $\zeta(n + m) = \zeta(n) \otimes \zeta(m)$ and the product as $\zeta(nm) = \zeta(n) \star \zeta(m)$. Numbers are encoded using a sum of modular vector representations, so $\zeta(n) = z_p(n) \otimes z_q(n) \otimes z_r(n) \ \otimes \ ...$, where $p, q, r, ...$ are positive coprime integers, and $z_s(n) = z_s(n \mod s)$. See \citeA{kymn_computing_2023}.

\subsubsection{Resonant decoding} As above, a resonator network is defined as $R(\zeta(n), P, Q, R, ...) = (z_p(n), z_q(n), z_r(n), ...)$ for moduli $p, q, r, ...$ Each set $S$ used in the decoding contains all the possible values of the corresponding function $z_s$, so $S = \{z_s(1), ..., z_s(s)\}$, since $z_s(n) = z_s(n \mod s)$. When a numeric representation is decoded into its modular constituents, the exact value $n$ can be decoded. Given we decode the tuple $(z_p(n \mod p), z_q(n \mod q), z_r(n \mod r), ...)$, we may further infer that the tuple of natural numbers $(n \mod p, \; n \mod q, \; n \mod r, ...)$ identifies the encoded number $n$,  That tuple uniquely encodes $n$ up to the least common multiple of $p, q, r, ...$ See \citeA{frady_resonator_2020}.

\subsection{Certain holographic declarations}

\citeA{Kelly2020} describes Holographic Declarative Memory (HDM), a \emph{declarative memory} module for the ACT-R cognitive architecture \cite{Anderson1993} that uses a VSA (specifically, holographic reduced representations; \citeNP{Plate2003}) to encode \emph{memory chunks}. In ACT-R, a chunk is a data structure that can be held in memory. In ACT-R, declarative memory contains both semantic and episodic memory, and, when \emph{probed} with the appropriate cue, will \emph{recall} whatever \emph{chunk} it stores that is most similar to the cue.

There are two types of information stored in an ACT-R chunk. They are \emph{sequential} information, and \emph{slot-value} information. Each kind of information consists of \emph{symbols} in the sense intended by \citeA{Newell1980}; namely, they \emph{refer to} some \emph{meaning} (i.e., another object of cognition), and that having a symbol confers \emph{distal access} to whatever meaning it symbolizes. \textit{Sequential} information is coded in a list-like format, where the symbols are ordered, and the position of each symbol matters. In the \emph{slot-value} format, symbols are stored in named, unordered \emph{slots}, and the chunk can be decomposed by retrieving symbols from it in some known slot. We are only interested in the slot-value encoding.

In HDM, each slot of a chunk has a permutation associated with it, $\mathbf{P}_{\mathrm{slot}}$. Objects of the same kind are stored in chunks that contain the same slots, so, if one were representing shapes of various colours and sizes, one might have chunks like \verb|(shape:circle colour:red size:large)| or \verb|(shape:square colour:blue size:small)|. A special placeholder value $\Phi$ is stored and held constant across all chunks. HDM is interested in the semantic content of values, for use in retrieval from declarative memory. As declarative memory is cued with chunks similar to what was stored, and probing it yields a chunk similar to the probe, \emph{values} are thought of as answers to the question ``what goes in the empty spot of my incomplete chunk?'' Given one is storing information about redness, instead of storing information about the value \emph{red} directly, HDM will store information about the context in which \textit{red} appears, and use a placeholder to stand in for \emph{red}. That is, supposing one has a large red circle in working memory, and one is storing information about redness, one stores $\mathbf{q}_{\mathrm{red}} = \mathbf{P}_{\mathrm{colour}}(\Phi) \otimes \mathbf{P}_{\mathrm{shape}}(v_{\mathrm{circle}}) \otimes \mathbf{P}_{\mathrm{size}}(v_{\mathrm{large}})$. The resulting stored value corresponding to the colour red is just the sum of all the stored contexts in which red occurs.

Because we can think of $\mathbf{q}_{\mathrm{red}}$ as a question to which the colour red is an answer (i.e., an incomplete chunk where red fills the colour slot), a chunk may be constructed as $\mathbf{c}(\mathrm{large \; red \; square}) = \mathbf{P}_{\mathrm{colour}}(v_{\mathrm{red}}) \otimes \mathbf{P}_{\mathrm{shape}}(v_{\mathrm{circle}}) \otimes \mathbf{P}_{\mathrm{size}}(v_{\mathrm{large}})$, and, should one be interested in describing a neural network that is able to complete type signatures given partial information, one can train a network to associate $v_{\mathrm{red}}$ with a distribution of questions like $\mathbf{q}_{\mathrm{red}}$. For our purposes, we are more interested in chunks $\mathbf{c}$. In the simple scheme presented by \citeauthor{Kelly2020} for HDM, chunks that are alike in structure and content will be spatially nearby. However, the HDM scheme is slightly too constraining, as chunks that are alike in structure, and similar in content, but unalike in one value, will be very dissimilar. Accordingly, we iterate on the HDM chunk representation by following a BEAGLE-like formula \cite{Jones2007}, and representing a chunk as the sum of the products of all subsets the chunk. In the preceding example, with $\mathcal{C} = \mathcal{P}(\{\mathrm{size:large\,,\; shape:circle\,,\; colour:red}\})$,  $\mathbf{c}(\mathrm{size:large \quad colour:red \quad shape:square}) = \sum_{c \in \mathcal{C}} \bigotimes_{\mathrm{slot:value}\in c} \mathbf{P}_{\mathrm{slot}}(v_{value})$, where $\mathcal{P}$ denotes a powerset. Chunks are normalized to a magnitude of 1. Because the unary subsets of a chunk are encoded in its representation, a value may be decoded from a chunk $\mathbf{c}$ by $\mathcal{M}(\mathbf{P}_{slot}^{-1}(\mathbf{c}))$, provided values are stored in the memory system.

\section{VSLRLLFPL, or, Doug}

\citeA{Flanagan2024} provide a method for encoding any arbitrary syntax into a VSA using
traditional \emph{role-filler pairs} commonly used in both symbolic \cite{Ritter2019actr} and vector-symbolic systems \cite{Plate2003,Smolensky1990}. Therefore, in order
to capture the polynomial time type system within the VSA, we propose the following encoding.

\begin{definition}[Doug Types]\label{doug-types}

    For the following definition symbols marked in \textbf{bold} will denote vector-symbols
    of a sufficiently high dimensionality $D$ sampled according to the dogma of the chosen VSA, except $\mathbf{c}$, which denotes the chunk constructor. The \emph{tags}
    of the encoding will be $\mathbf{Boolean}$, $\mathbf{Map}$, $\mathbf{Tuple}$,
    $\mathbf{List}$, $\mathbf{Bang}$, and $\mathbf{Credit}$.

    Recalling the encoding scheme we derived from HDM above, a chunk of slot-value pairs is encoded as,  $\mathbf{c}(\mathrm{slot_1}:\mathbf{value_1} \quad \mathrm{slot_2}:\mathbf{value_2} ...) = \sum_{c \in \mathcal{C}} \bigotimes_{\mathrm{slot:value}\in c} \mathbf{P}_{\mathrm{slot}}(\mathbf{value})$, where $\mathcal{C} = \mathcal{P}(\{\mathrm{slot_1:value_1\,,\; slot_2:value_2\,,} ...)$, we construct the following types inductively.

    For each type in Def.~\autoref{def}, we proceed step-wise with an encoding function:
    \begin{align*}
        \text{boolean}(n) &:= \mathbf{c}(\mathrm{kind}:\mathbf{Boolean} \quad \mathrm{type}:\mathbf{B} \quad \mathrm{level}: n), \\
        \text{map}(d, c) &:= \mathbf{c}(\mathrm{kind} : \mathbf{Map}  \\ & \qquad \; \mathrm{type}: \mathbf{c}(\mathrm{dom}:d \quad \mathrm{codom}: c), \quad \mathrm{level}:n), \\
        \text{tuple}(l, r) &:= \mathbf{c}(\mathrm{kind} : \mathbf{Tuple} \\ & \qquad \; \mathrm{type}: \mathbf{c}(\mathrm{left}:l \quad \mathrm{right}: r), \quad \mathrm{level}:n), \\
        \text{list}(n, s) &:= \mathbf{c}(\mathrm{kind} : \mathbf{List} \quad \mathrm{type}: s, \quad \mathrm{level}:n), \\
        \text{bang}(n, s) &:= \mathbf{c}(\mathrm{kind} : \mathbf{Bang} \quad \mathrm{type}: s, \quad \mathrm{level}:n), \\
        \text{credit}(n) &:= \mathbf{c}(\mathrm{kind} : \mathbf{Credit} \quad \mathrm{type}: \mathbf{D}, \quad \mathrm{level}:n).
    \end{align*}
    Where $n$ uniformly denotes a residue natural number.
\end{definition}
The encoding allows for (1) structuring the different sub-elements of the types in 
a compositional manner, (2) querying a representation for a specific sub-element, and (3)
a notion of \emph{similarity} between two types.

We encode the constants found in Def.~\autoref{const} as follows:
\begin{definition}[Doug Constants]
    We maintain the same convention as Def.~\autoref{doug-types} for denoting
    vector-symbols. Let us have tag vector-symbols $\mathbf{TT}$, $\mathbf{FF}$,
    $\mathbf{Case_{bool}}$, $\mathbf{Case_{list}}$, $\mathbf{Cons}$,
    $\mathbf{Nil}$, $\mathbf{Dollar}$, $\mathbf{Pair}$, and $\mathbf{Proj}$.

    We proceed step-wise for each item in Def.~\autoref{const},
    \begin{align*}
        \text{tt}(n) &:= \mathbf{TT} + (\mathbf{level} \otimes n), \\
        \text{ff}(n) &:= \mathbf{FF} + (\mathbf{level} \otimes n), \\
        \text{case}_\text{bool}(n, s) &:= \mathbf{Case_{bool}} + (\mathbf{level} \otimes n) + (\mathbf{type} \otimes s), \\
        \text{case}_\text{list}(n, t, s) &:= \mathbf{Case_{list}} + (\mathbf{level} \otimes n) + (\mathbf{from} \otimes t) + (\mathbf{to} \otimes s),\\
        \text{cons}(n, t)&:= \mathbf{Cons} + (\mathbf{level} \otimes n) + (\mathbf{type} \otimes t), \\
        \text{nil}(n, t)&:= \mathbf{Nil} + (\mathbf{level} \otimes n) + (\mathbf{type} \otimes t), \\
        \text{dollar}(n) &:= \mathbf{Dollar} + (\mathbf{level} \otimes n), \\
        \text{pair}(n, l, r) &:= \mathbf{Pair} + (\mathbf{level} \otimes n) + (\mathbf{left} \otimes l) + (\mathbf{right} \otimes r), \\
        \text{proj}(n, s)&:= \mathbf{Proj}+ (\mathbf{level}\otimes n) + (\mathbf{type} \otimes s),
    \end{align*}
    where again $n$ is some natural number encoding.
\end{definition}
Constant terms of the language represent \textit{constructors} and \textit{destructors} of types, which are ways we 
can express type introduction and elimination \cite[pg. 27]{hottbook}.

Types and constants do not make up the whole language: we must also have a way of encoding arbitrary expressions.
\begin{definition}[Doug Terms]
    We adopt the same vector-symbolic conventions above and sample vectors of the same  
    dimensionality. Following Def.~\autoref{terms}, let the tag symbols be $\mathbf{annotation}$,
    $\mathbf{Const}$, $\mathbf{Lambda}$, $\mathbf{App}$, $\mathbf{Box}$, $\mathbf{Brackets}$,
    and $\mathbf{Sub}$.

    Step-wise, the encoding of terms is as follows:
    \begin{align*}
        \text{annotation}(x, \tau) &:= \mathbf{Annotation} + (\mathbf{var} \otimes x) + (\mathbf{type} \otimes \tau), \\
        \text{const}(c) &:= \mathbf{Const} + (\mathbf{val} \otimes c),
    \end{align*}
    \begin{align*}
        \text{lambda}(x, \tau, t) &:= \mathbf{Lambda} + (\mathbf{var} \otimes x) + 
        (\mathbf{type} \otimes \tau) \\
        &+ (\mathbf{body} \otimes t), \\
        \text{app}(t, s) &:= \mathbf{App} + (\mathbf{rator} \otimes t) + (\mathbf{rand} \otimes s), \\
        \text{box}(n, x, s, t) &:= \mathbf{Box} + (\mathbf{let} \otimes x) + 
        (\mathbf{this} \otimes s) \\ &+ (\mathbf{that} \otimes t) + (\mathbf{level} \otimes n), \\
        \text{brace}(n, t) &:= \mathbf{Brace} + (\mathbf{level} \otimes n) + (\mathbf{term} \otimes t), \\
        \text{sub}(s, n, x_1, x_2, t) &:= \mathbf{Sub}+ (\mathbf{this} \otimes t) + 
        (\mathbf{that} \otimes s) \\ &+ (\mathbf{level} \otimes n) + (\mathbf{x1} \otimes x_1) 
        + (\mathbf{x2} \otimes x_2),
    \end{align*}
    where $n$ is some natural number encoded in HRR.
\end{definition}

\section{Discussion}

Doug allows us to encode types over a vector space. We view types as a sort of \emph{constraint} on program synthesis; given a type, one narrows a search space of possible programs. If a type is chosen reasonably, and the type system is sufficiently constraining, the search space may be constrained to such an extent that searching for a program satisfying some goal can be done in polynomial, not exponential time. Finding an optimal program, we expect, will remain computationally hard. But finding any program that satisfies a goal, and furthermore, constraining a search to only consider programs of constant behaviour should not be, as, during skill acquisition, humans tend to do so with ease: human skill acquisition is not doubly exponential, connoting exhaustive search both for more efficient programs and programs of the correct behaviour (as suggested by \citeNP[ch. 1]{Hutter2000}), but singly-exponential, connoting a hard search for more efficient procedures, but an easier search for correct behaviour \cite{Heathcote2000}.

\citeA{Dehaene2022} found that the neural representations for at least some simple skills seem to have the information content of optimal programs. There are two suggestions that follow from their finding: First, that human mental representations must be expressive enough to store arbitrary programs, at least up to some maximum complexity permitted by memory capacity. Second, humans are fairly good at finding optimal representations, up to the limits of what is tractable. This result is unsurprising: \citeauthor{Hutter2005} found that, given an optimal representation of the world, rational goal-directed behaviour is simple to achieve. Humans tend to behave rationally, given the resources to do so. Under the commitments of the common model of cognition \cite{Laird2017}, humans are ``boundedly rational'', or, one might say, \emph{rational, up to the limits of what is tractable}, and that \emph{the degree of rational behaviour one can achieve is a skill issue}.

\citeA{Flanagan2024} argued that VSAs provide the machinery necessary to interpret brain states as syntactically structured representations; in other words, if humans solve problems by generating \textit{skill programs} that \emph{may be optimal} for some tasks, human brain states must encode programs. Furthermore, humans must learn to generate those programs efficiently, and, in order to do so, their search for those programs must be constrained to a subset of programs that may be useful. Those constraints must be sufficiently strict to radically accelerate program synthesis relative to brute-force search, and must \textit{also} be interpretable as brain states, and must be learnable, as humans somehow acquire knowledge of the constraints appropriate to novel problems.

Doug is the first step in describing \textit{learnable, provably strong constraints} that might limit the complexity of program synthesis. By constraining a type system to express only programs that run in polynomial time, programs typed with Doug may express only tractable solutions to given problems. However, it remains to be shown which, if any, type systems can make program synthesis polynomial, and whether learning over those types does not take so long as to cancel any benefits gleaned from constraint. Nevertheless, Doug captures important intuitions about what human skill acquisition should be like, and provides a methodology by which future type systems that really constrain program synthesis like humans can be devised.

\bibliographystyle{apacite}

\setlength{\bibleftmargin}{.125in}
\setlength{\bibindent}{-\bibleftmargin}

\bibliography{references}

\end{document}